\documentclass{article}
\usepackage{spconf,amsmath,graphicx,hyperref}
\usepackage{amssymb}
\usepackage{booktabs}
\usepackage{multirow}
\usepackage{tabularx}
\usepackage{makecell}
\usepackage{cite}
\usepackage{amsfonts}
\usepackage{algorithmic}
\usepackage{textcomp}
\usepackage{xcolor}
\usepackage{array}
\usepackage{bbding}

\title{MMRQA: Signal-Enhanced Multimodal Large Language Models for MRI Quality Assessment}

\name{%
\begin{tabular}{@{}c@{}}
Fankai Jia$^{1,2,*}$ \qquad 
Daisong Gan$^{1,2,*}$ \qquad 
Zhe Zhang$^{1,2}$ \qquad 
Zhaochi Wen$^{1,3}$ \\
Chenchen Dan$^{4}$ \qquad 
Dong Liang$^{1,2,\dag}$ \qquad 
Haifeng Wang$^{1,2,\dag}$
\end{tabular}
\thanks{$^*$Authors contributed equally to this work. $^\dag$Corresponding authors.}
}

\address{
$^1$Shenzhen Institutes of Advanced Technology, Chinese Academy of Sciences, Shenzhen, China \\
$^2$University of Chinese Academy of Sciences, Beijing, China \\
$^3$ShanghaiTech University, Shanghai, 201210, China \\
$^4$Southern University of Science and Technology, Shenzhen, 518020, China \\
\{jiafankai24, gandaisong23\}@mails.ucas.ac.cn
}

\begin{document}
\ninept
\maketitle

\begin{abstract}
Magnetic resonance imaging (MRI) quality assessment is crucial for clinical decision-making, yet remains challenging due to data scarcity and protocol variability. Traditional approaches face fundamental trade-offs: signal-based methods like MRIQC provide quantitative metrics but lack semantic understanding, while deep learning approaches achieve high accuracy but sacrifice interpretability. To address these limitations, we introduce the Multimodal MRI Quality Assessment (MMRQA) framework, pioneering the integration of multimodal large language models (MLLMs) with acquisition-aware signal processing. MMRQA combines three key innovations: robust metric extraction via MRQy augmented with simulated artifacts, structured transformation of metrics into question-answer pairs using Qwen, and parameter-efficient fusion through Low-Rank Adaptation (LoRA) of LLaVA-OneVision. Evaluated on MR-ART, FastMRI, and MyConnectome benchmarks, MMRQA achieves state-of-the-art performance with strong zero-shot generalization, as validated by comprehensive ablation studies. By bridging quantitative analysis with semantic reasoning, our framework generates clinically interpretable outputs that enhance quality control in dynamic medical settings.
\end{abstract}

\section{Introduction}
Magnetic resonance imaging (MRI) interpretation faces substantial clinical challenges, with diagnostic errors influenced by cognitive biases and system factors leading to variable error rates across centers~\cite{Lee2013,Herzog2017}. The complexity of MRI acquisition, prone to artifacts from motion, inhomogeneity, and signal interference, necessitates robust, automated quality control to support accurate decision-making. Traditional manual assessment suffers from inter-rater variability and scalability limitations for large datasets~\cite{esteban2017mriqc}. These fundamental challenges have spurred automated methods, progressively evolving from unsupervised metric extraction to AI-driven semantic analysis.

\begin{figure}[t]
    \centering
    \includegraphics[width=.9\linewidth]{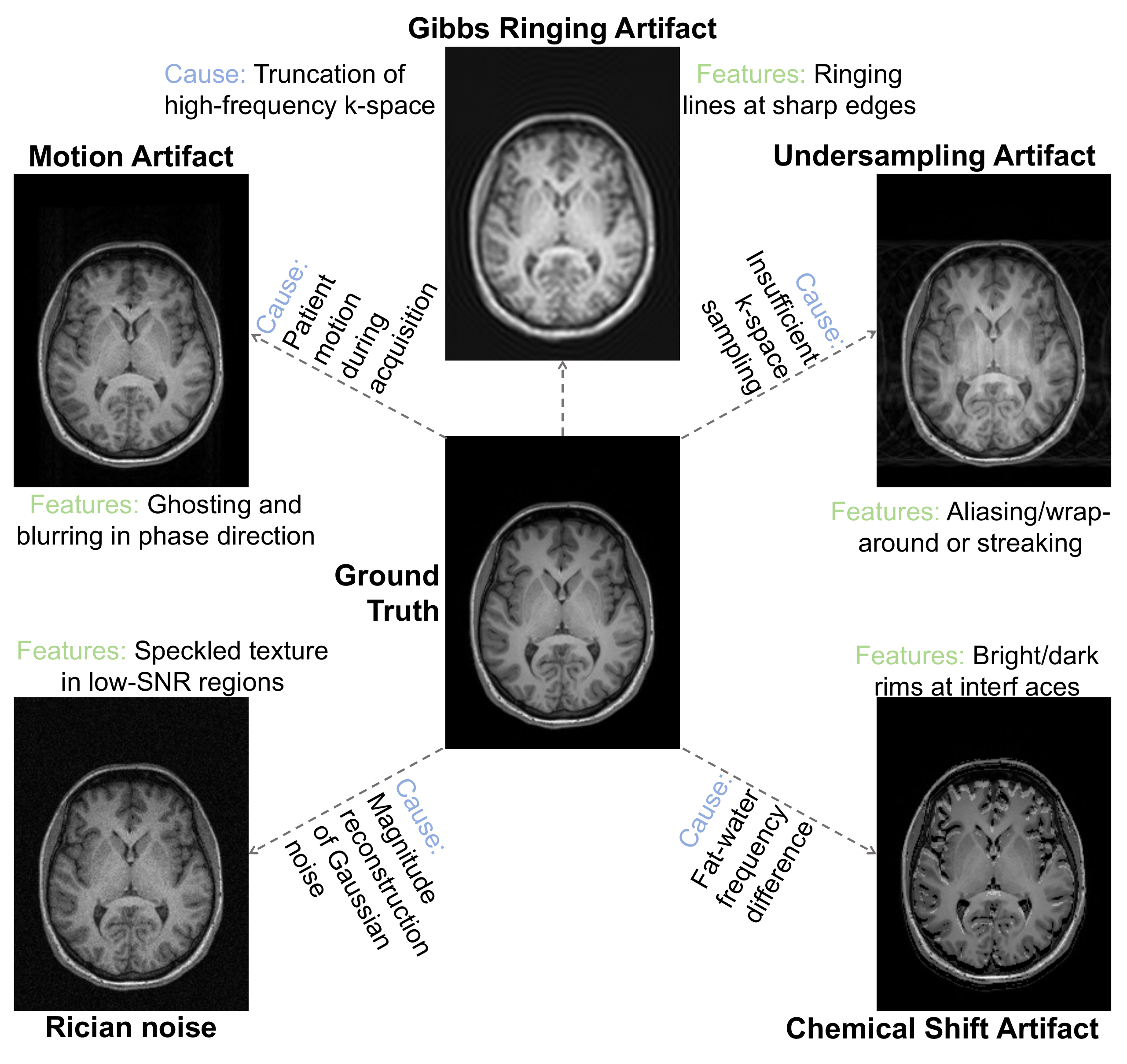}
    \caption{Taxonomy of MRI artifacts: Mapping acquisition causes to visual manifestations in T1w scans. The radial layout from a central reference highlights signal degradation patterns, motivating the need for multimodal assessment strategies.}
    \label{fig:fig1}
\end{figure}

\begin{figure*}[ht]
    \centering
    \includegraphics[width=.8\textwidth]{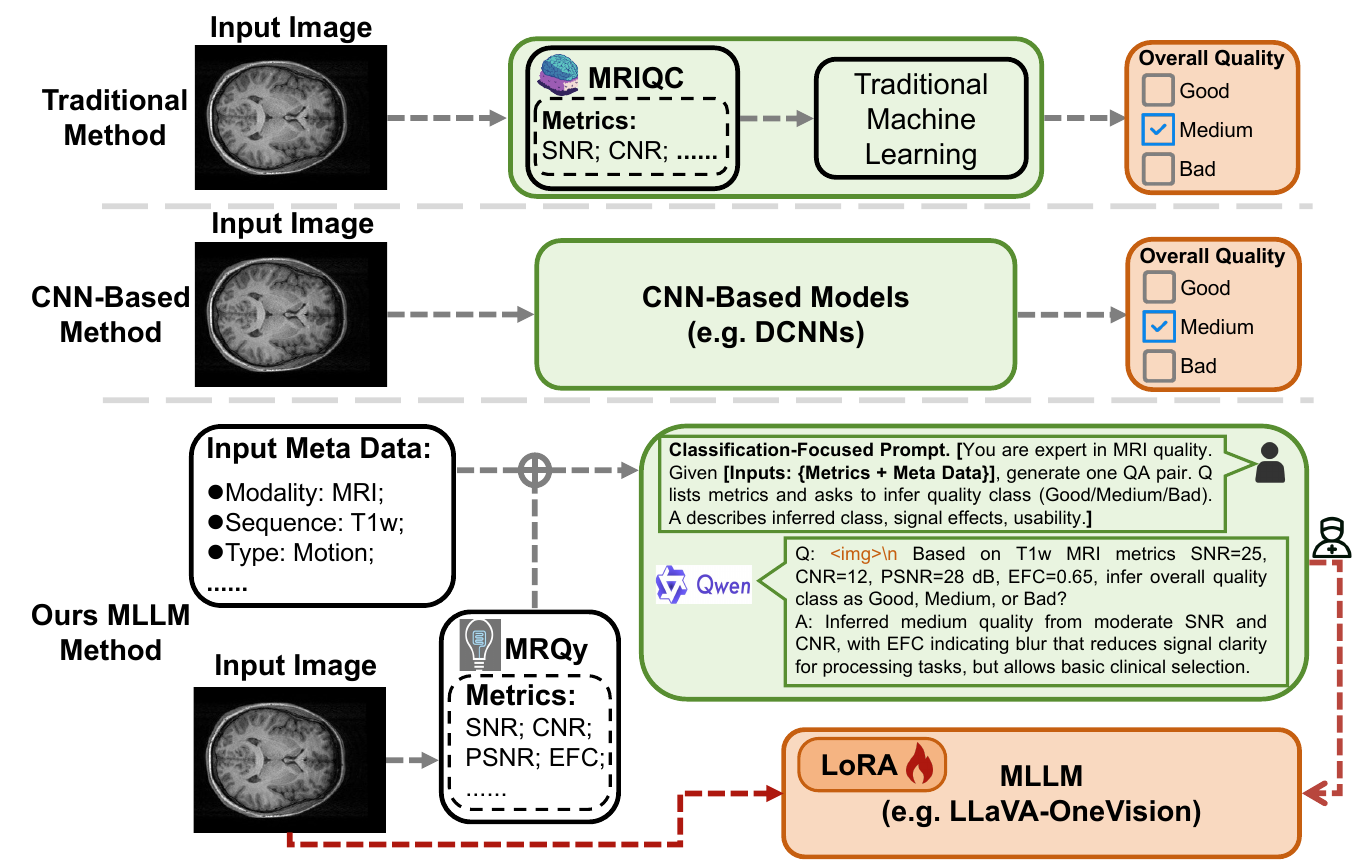}
    \caption{Comparative overview of the MMRQA pipeline. Traditional paths employ MRQy/MRIQC for metric extraction or CNNs (e.g., DCNNs) for visual classification, while MMRQA integrates Qwen-generated QA pairs with LoRA fine-tuning of LLaVA-OneVision for multimodal metric-visual reasoning and inference.}
    \label{fig:fig2}
\end{figure*}

Existing automated approaches exhibit three fundamental trade-offs in MRI quality assessment, reflecting the field's progressive yet fragmented evolution. Signal-based metrics prioritize quantitative objectivity but sacrifice semantic understanding: MRIQC~\cite{esteban2017mriqc} and MRQy~\cite{sadri2020mrqy} extract comprehensive indicators (SNR, CNR, EFC) for cross-site standardization and batch effect detection, while FetMRQC~\cite{Sanchez2024} extends this to fetal-specific ensembles robust to motion, yet these methods fail to capture complex artifact patterns requiring contextual interpretation. This limitation motivated visual deep learning approaches that excel at pattern recognition but compromise explainability: ensemble DCNNs~\cite{Sujit2019}, image rulers with multi-task networks~\cite{Lei2021}, attention-based prototypes~\cite{garcia2024brainqcnet}, 3D pipelines for diffusion degradations~\cite{ahmad2023-3dqcnet}, data ramping with uncertainty quantification~\cite{pizarro2023deep}, and multi-level fusions for no-reference assessment~\cite{stepien2023noreference} enable nuanced detection and generalization, yet their black-box nature offers limited actionable clinical guidance. Recent multimodal attempts~\cite{miao2025ultrasound,you2024descriptive} begin incorporating language for interpretability through prompt-based evaluation and vision-language descriptive assessment, but remain confined to superficial semantic-visual alignment without true fusion. Each paradigm optimizes one aspect (objectivity, accuracy, or interpretability) while compromising others, underscoring the need for unified frameworks that integrate all three to achieve robust, clinically viable quality assessment.

To address these limitations, we introduce the Multimodal MRI Quality Assessment (MMRQA) framework, pioneering the first application of MLLMs to MRI quality evaluation and marking a transformative shift in the field. By harnessing MLLMs' semantic fusion and contextual reasoning capabilities, MMRQA enables robust quality assessment through three interconnected innovations that exploit their unique ability to integrate textual priors\textemdash unlike CNN-based methods limited to visual processing: (1) acquisition-aware signal characterization, extracting metrics like SNR and CNR via MRQy~\cite{sadri2020mrqy} and augmenting with simulated artifacts to enhance dataset robustness; (2) structured quality representation, converting numerical metrics into semantically rich question-answer (QA) pairs using Qwen~\cite{yang2024qwen2} for enhanced interpretable inference; and (3) parameter-efficient multimodal fusion, adapting LLaVA-OneVision~\cite{li2024llava} through Low-Rank Adaptation (LoRA)~\cite{hu2022lora} to seamlessly align visual and textual features. This integrated framework bridges quantitative signal processing with clinical decision-making, achieving superior robustness over traditional approaches while generating actionable, interpretable insights. Leveraging pretrained multimodal knowledge, MMRQA efficiently adapts to emerging artifact types with minimal labeled data, effectively tackling data scarcity in evolving clinical environments.

\section{METHODOLOGY}
\subsection{Problem Formulation and Quality-Aware Architecture}
Traditional MRI quality assessment methods, such as CNN-based classifiers and conventional machine learning approaches, rely on pixel-level features or predefined rules, limiting semantic interpretations of artifacts and explainable clinical decision-making. These methods often overlook integrating quantitative signal metrics with higher-level reasoning, reducing robustness across protocols and actionable insights. To address this, we introduce the Multimodal MRI Quality Assessment (MMRQA) framework, which leverages multimodal large language models (LLMs) for semantic fusion and contextual understanding, introducing a novel quality-aware architecture for MRI analysis. MMRQA models quality indicators as informative features in an integrated pipeline for robust classification and explainable reasoning. To our knowledge, this is the first systematic integration of MRI signal quality metrics with multimodal LLMs via structured QA transformation, advancing quality-aware medical image analysis beyond pixel-level processing to clinically interpretable outcomes.

We formulate the problem as a hierarchical fusion of visual data $X_v \in \mathbb{R}^{H \times W \times D}$ and metadata $X_m$, where signal metrics extracted from $X_v$ are transformed into semantic representations for multimodal inference:
\begin{equation}
p(Y | X_v, X_m) = f_{\text{fusion}}(g_{\text{LLM}}(h_{\text{q}}(X_v, X_m)), \phi(X_v)).
\label{eq:framework}
\end{equation}
Here, $h_{\text{q}}$ computes indicators like SNR and CNR using MRQy \cite{sadri2020mrqy}, $g_{\text{LLM}}$ generates QA pairs via Qwen \cite{yang2024qwen2}, $\phi$ encodes visuals through a frozen encoder, and $f_{\text{fusion}}$ performs LoRA-based adaptation \cite{hu2022lora}.
As shown in Fig.~\ref{fig:fig2}, the MMRQA pipeline includes three stages: acquisition-aware metric characterization, structured QA representation using frozen LLMs, and parameter-efficient multimodal adaptation. This design enables baseline comparisons and generalization across MRI protocols, bridging quantitative signal processing with semantic reasoning for enhanced clinical utility.

\subsection{Acquisition-Aware Signal Characterization}
In the first stage of MMRQA, we perform acquisition-aware signal characterization to extract robust quantitative metrics that ground the multimodal reasoning process. We employ MRQy~\cite{sadri2020mrqy}, an open-source tool for unsupervised MRI quality control, to derive 15 metrics from input volumes, including noise indicators like SNR variants, contrast measures like CNR, sharpness metrics like CPP, and artifact detectors like EFC and FBER. These metrics quantify signal integrity, tissue contrast, and artifact presence, forming a foundational layer for semantic transformation. For instance, one key metric, SNR, is computed as:
\begin{equation}
\text{SNR} = \frac{\mu_F}{\sigma_B},
\end{equation}
where $\mu_F$ is the mean foreground intensity and $\sigma_B$ the background noise standard deviation. To address data scarcity, we augment the dataset with simulated artifacts (e.g., motion, inhomogeneity), effectively bridging raw MRI data with interpretable quality representations to enhance the overall multimodal model performance in quality assessment. This characterization provides a quantitative foundation that is semantically enriched in the subsequent structured representation stage.

\subsection{Structured Quality Representation via Language Models}
Building upon the acquisition-aware metrics extracted in the previous stage of MMRQA, we employ the Qwen-Max model via its API \cite{yang2024qwen2} as a reasoning engine to transform these quantitative indicators into structured question-answer (QA) pairs. Since MRQy outputs consist solely of numerical image quality indicators, which are not in the natural language sentence format typically processed by large language models (LLMs), this conversion makes them more amenable to LLM processing while capturing semantic relationships essential for multi-task inference.
Modular prompts guide the generation of concise QA pairs across key tasks:
\begin{itemize}
    \item Classification: Infer quality levels (Good/Medium/Bad) from metrics, along with descriptions of signal effects and usability.
    \item Artifact: Deduce artifact types, causes (e.g., k-space disruption), and features (e.g., blur or ghosting).
    \item Analysis: Evaluate usability for downstream processing and suggest optimizations (e.g., PROPELLER sequences).
\end{itemize}
The generated QA pairs undergo expert review to ensure self-containment and clinical relevance. This transformation creates a shared semantic space between quantitative metrics and visual features, facilitating effective multimodal adaptation while preserving interpretability. These QA pairs thus enable efficient alignment in the subsequent parameter-efficient fusion stage of MMRQA.

\subsection{Parameter-Efficient Multimodal Fusion}
In the final stage of MMRQA, we perform parameter-efficient multimodal fusion to align the structured QA representations with visual features for comprehensive quality inference. We adopt LLaVA-OneVision \cite{li2024llava}, which combines SigLIP vision encoding with Qwen2 \cite{yang2024qwen2}, enhanced via parameter-efficient fine-tuning for clinical feasibility. We apply LoRA \cite{hu2022lora} to decompose weight updates, reducing trainable parameters. The update is:
\begin{equation}
W' = W_0 + \frac{\alpha}{r}BA,
\end{equation}
where \(W_0\) is the original weight matrix, \(B\) and \(A\) are low-rank matrices, \(r\) is the rank, and \(\alpha\) is the scaling factor.
This fusion aligns QA-transformed metrics with visual features, enabling chained reasoning across tasks while preserving pretrained knowledge. By leveraging LoRA's efficiency, MMRQA supports rapid deployment in resource-constrained clinical settings, effectively bridging quantitative signal processing with semantic outputs for robust and interpretable MRI quality assessment.

\begin{figure}[t]
    \centering
    \includegraphics[width=.9\columnwidth]{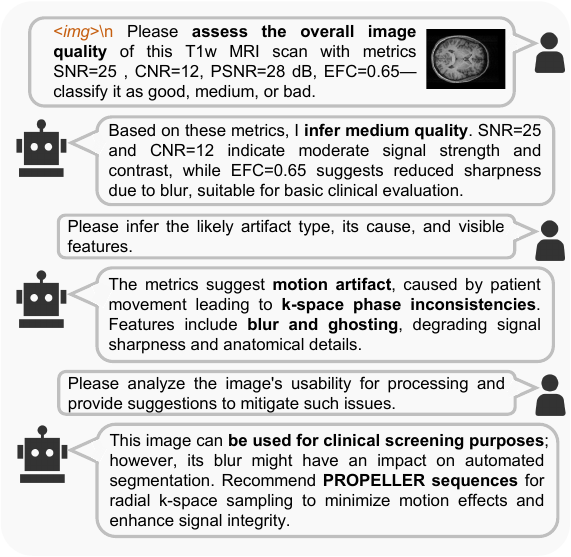}
    \caption{Multi-round quality assessment dialogue demonstrating MMRQA's progression from signal metrics to artifact identification and clinical recommendations.}
    \label{fig:fig3}
\end{figure}

\section{Experiments}

\begin{table*}[ht]
\centering
\caption{Performance comparison across three MRI quality assessment benchmarks. GPT-5 evaluates generated textual descriptions; baseline methods produce only numerical scores. Best results in bold, second-best underlined.}
\label{tab:main_results}
\setlength{\tabcolsep}{3pt}
\begin{tabular}{l|ccc|ccc|ccc}
\toprule
\multirow{2}{*}{Method} & \multicolumn{3}{c|}{MR-ART} & \multicolumn{3}{c|}{FastMRI} & \multicolumn{3}{c}{MyConnectome} \\ 
& Acc$\uparrow$ & F1$\uparrow$ & GPT-5$\uparrow$ & Acc$\uparrow$ & F1$\uparrow$ & GPT-5$\uparrow$ & Acc$\uparrow$ & F1$\uparrow$ & GPT-5$\uparrow$ \\
\midrule
MRIQC~\cite{esteban2017mriqc} & 71.2 & 68.4 & N/A & 68.5 & 65.2 & N/A & 46.3 & 43.1 & N/A \\
FetMRQC~\cite{Sanchez2024} & 74.5 & 71.3 & N/A & 71.8 & 68.9 & N/A & 49.7 & 46.5 & N/A \\
\midrule
3P-Ensemble~\cite{Sujit2019} & 79.1 & 76.7 & N/A & 75.4 & 73.1 & N/A & 52.6 & 50.1 & N/A \\
IRQA~\cite{Lei2021} & 78.3& 75.6 & N/A & \underline{78.2} & 72.5 & N/A & 54.0 & 51.2 & N/A \\
DRAM-Net~\cite{pizarro2023deep} & 77.8 & 74.9 & N/A & 74.9 & 73.6 & N/A & \underline{56.3} & 49.4 & N/A \\
MLMF-MRIQA~\cite{stepien2023noreference} & \underline{80.4} & \underline{78.4} & N/A & 77.6 & \underline{74.1} & N/A & 55.2 & \underline{53.5} & N/A \\
\midrule
\textbf{MMRQA (Ours)} & \textbf{88.1} & \textbf{85.7} & \textbf{85.4} & \textbf{85.1} & \textbf{82.8} & \textbf{83.7} & \textbf{72.2} & \textbf{66.7} & \textbf{68.9} \\
\bottomrule
\end{tabular}
\end{table*}

\subsection{Experimental Setup}

Our MMRQA framework employs LLaVA-OneVision~\cite{li2024llava}, 
an architecture combining SigLIP vision encoding with the Qwen2 
language backbone. We leverage the ms-swift framework~\cite{zhao2025swift} for parameter-efficient fine-tuning through LoRA, with rank $r=16$ and scaling factor $\alpha=16$. Training is conducted on 2 NVIDIA A6000 GPUs using Adam optimizer (learning rate=0.0001, $\beta_1=0.9$, $\beta_2=0.95$) with batch size 8. A single epoch suffices due to structured knowledge incorporation from MRQy metrics~\cite{sadri2020mrqy}.

\subsection{Datasets and Evaluation Protocol}
Our evaluation employs three complementary benchmarks to validate MMRQA's performance across controlled assessment, data augmentation, and generalization. MR-ART~\cite{narai2022mrart}, with 148 subjects featuring matched motion-corrupted and clean brain scans, enables controlled evaluation; we reserve 30\% as a held-out test set without training involvement, using the remainder for training. To mitigate data scarcity in medical imaging, we combine this with FastMRI~\cite{knoll2020fastmri}, augmented through simulated artifacts including motion blur, k-space ghosting, and aliasing, thereby exposing the model to diverse degradation patterns without extensive manual labeling. MyConnectome~\cite{poldrack2015longterm}, comprising 104 longitudinal sessions, serves as a zero-shot benchmark for temporal generalization, assessing adaptation to protocol variations and scanner changes without retraining.

\subsection{Evaluation Metrics}
Our evaluation combines quantitative metrics with interpretive assessment to capture both algorithmic accuracy and clinical utility. We employ Accuracy and macro F1-score for three-tier quality classification, ensuring balanced performance across classes. To evaluate MMRQA's generated quality descriptions, we adapt GPT-5 evaluation~\cite{openai2025gpt5} following descriptive paradigms~\cite{you2024descriptive}, scoring on a 0-100 scale for artifact identification accuracy (type, cause, features) and diagnostic relevance (actionability and alignment with clinical needs). GPT-5 is prompted as: ``Assess the following MRI quality description [generated text] on a 0-100 scale for artifact identification and diagnostic relevance.'' Scoring aggregates multiple runs for consistency, with higher values indicating stronger clinical alignment and addressing limitations in purely numerical approaches.

\subsection{Results and Analysis}
Table~\ref{tab:main_results} demonstrates MMRQA's state-of-the-art performance across all benchmarks. The framework ubstantially surpasses existing approaches, achieving consistent improvements over both traditional feature-based methods and deep learning baselines. Notably, MMRQA maintains robust performance on zero-shot MyConnectome evaluation, validating generalization across temporal variations and acquisition protocols. The GPT-5 scores confirm MMRQA's unique capability to generate clinically interpretable descriptions beyond numerical assessment, thereby addressing the critical communication gap between automated systems and clinical practice. This multimodal approach successfully bridges quantitative signal analysis with semantic understanding.

\begin{table}[t]
\centering
\caption{Ablation study on MR-ART. QA percentages indicate proportion of question-answer pairs utilized.}
\label{tab:ablation}
\begin{tabular}{l|ccc}
\toprule
Variant & Acc$\uparrow$ & F1$\uparrow$ & GPT-5$\uparrow$ \\
\midrule
Full Model & \textbf{88.1} & \textbf{85.7} & \textbf{85.4} \\
No MRQy & 82.3 & 80.7 & 72.1 \\
With 67\% QA & 87.2 & 84.9 & 84.2 \\
With 33\% QA & 85.7 & 83.6 & 82.8 \\
\bottomrule
\end{tabular}
\end{table}

\subsection{Ablation Study}
Table~\ref{tab:ablation} reveals architectural component contributions. Removing MRQy metrics substantially degrades performance, confirming quantitative characterization as essential grounding. Progressive QA reduction shows graceful degradation: 67\% sampling maintains near-optimal performance, while 33\% sampling still achieves competitive results. These findings validate the synergistic benefit of integrating acquisition-aware signal processing with structured language transformation.

\section{CONCLUSION}
To address the fundamental disconnect between automated MRI assessment and clinical interpretation, we presented MMRQA, pioneering the application of MLLMs to MRI medical image quality evaluation. By synergistically integrating quantitative signal analysis with semantic reasoning capabilities, our framework transcends traditional numerical scoring to deliver contextually rich, clinically actionable insights. Experimental validation confirms MMRQA's robust performance across diverse acquisition protocols and its unique ability to generate interpretable quality narratives. Beyond quality assessment, MMRQA's human-aligned evaluation paradigm offers promising applications for reconstruction algorithms, where conventional metrics like PSNR correlate poorly with radiological utility. This approach could guide the development of reconstruction methods that optimize perceptual quality rather than merely minimizing pixel-wise errors, ultimately enhancing diagnostic confidence in clinical practice.


\vfill
\textbf{Acknowledgements.} The work is supported by Guangdong Provincial Key Laboratory of Multimodality Non-Invasive Brain-Computer Interfaces (Grant no.2024B1212010010).

\textbf{Disclosure of Competing Interests.} The authors declare no competing interests relevant to the content of this article.
\pagebreak

\bibliographystyle{IEEEbib}
\bibliography{strings,refs}

\end{document}